\documentclass[runningheads]{llncs}
\usepackage[T1]{fontenc}
\usepackage{graphicx}
\begin{document}
\title{Realistic pedestrian behaviour  in the CARLA simulator using VR and mocap} 
%
%
\author{Sergio Martín Serrano\inst{1}\orcidID{0000-0002-8029-1973} \and
David Fernández Llorca\inst{1,2}\orcidID{0000-0003-2433-7110} \and
Iván García Daza\inst{1}\orcidID{0000-0001-8940-6434} \and
Miguel Ángel Sotelo\inst{1}\orcidID{0000-0001-8809-2103}}
\authorrunning{Martín Serrano, S., Fernández Llorca, D., García Daza, I., Sotelo, M. A.}
%
\institute{Computer Engineering Department, University of Alcalá, Alcalá de Henares, Spain \and
European Commission, Joint Research Centre (JRC), Seville} 


%
\maketitle              
\begin{abstract}
Simulations are gaining increasingly significance in the field of autonomous driving due to the demand for rapid prototyping and extensive testing. Employing physics-based simulation brings several benefits at an affordable cost, while mitigating potential risks to prototypes, drivers, and vulnerable road users. However, there exit two primary limitations. Firstly, the \emph{reality gap} which refers to the disparity between reality and simulation and prevents the simulated autonomous driving systems from having the same performance in the real world. Secondly, the lack of empirical understanding regarding the \emph{behavior of real agents}, such as backup drivers or passengers, as well as other road users such as vehicles, pedestrians, or cyclists. Agent simulation is commonly implemented through deterministic or randomized probabilistic pre-programmed models, or generated from real-world data; but it fails to accurately represent the behaviors adopted by real agents while interacting within a specific simulated scenario. This paper extends the description of our proposed framework to enable real-time interaction between real agents and simulated environments, by means immersive virtual reality and human motion capture systems within the CARLA simulator for autonomous driving. We have designed a set of usability examples that allow the analysis of the interactions between real pedestrians and simulated autonomous vehicles and we provide a first measure of the user's sensation of \emph{presence} in the virtual environment.

\end{abstract}
\section{Introduction}
The rise in the use of simulators in the context of autonomous driving is mainly due to the need for prototyping and exhaustive validation, since the tests of autonomous systems directly on real scenarios alone are not capable of providing sufficient evidence that prove its safety \cite{ref_Kalra2016}.  
There is some initial consensus that future testing approaches should be multisystem, including not only physical testing on proving grounds but also extensive use of simulators and real-world driving tests \cite{ref_Llorca2023}.
With simulators we can generate large amounts of data, including edge cases, and enrich training and testing with a specific control over all variables under study (e.g., street layout, lighting conditions, traffic scenarios). Furthermore, the generated data can be also annotated by design including semantic information. This is particularly interesting when testing predictive systems \cite{ref_Izquierdo2022}.

However, one of the main challenges in developing autonomous driving simulators is the unrealistic nature of the data generated by simulated sensors and physical models. The well-known \emph{reality gap} leads to inaccuracies since the virtual world does not properly generalise all the variations and complexities of the real world \cite{ref_Stocco2022}, \cite{ref_GarciaDaza2022}. Additionally, despite there have been efforts to create lifelike artificial behaviors for other agents on the road (e.g., vehicles, pedestrians, cyclists), simulations are limited by a lack of empirical knowledge about their actual behavior. As a result, this gap affects both behavior and movement prediction as well as human-vehicle communication and interaction \cite{ref_Eady2019}.

\begin{figure*}
  \centering
  \includegraphics[width=\linewidth]{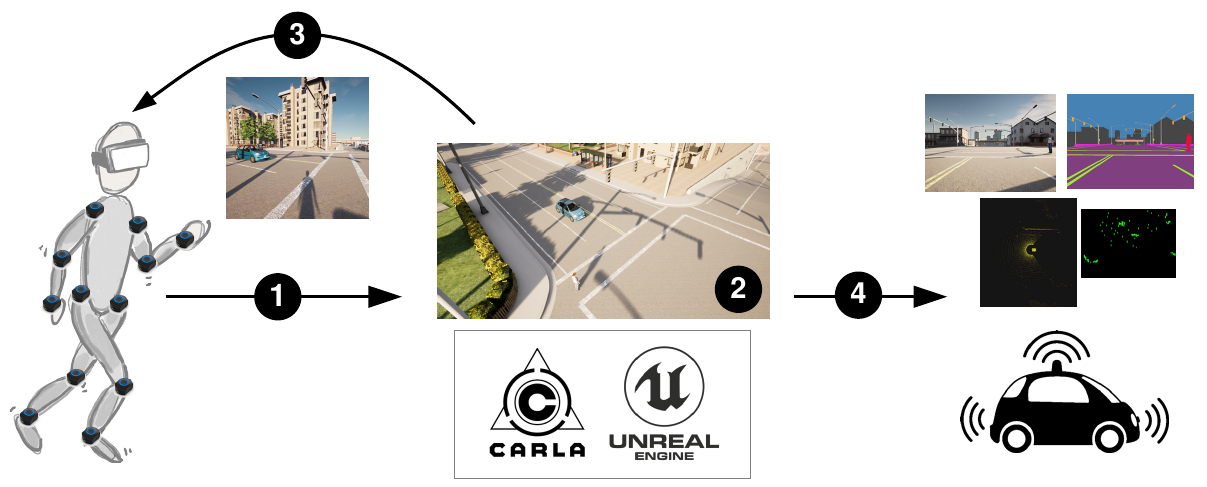}
  \caption{Overview of the presented approach. Adapted from \cite{ref_CARLA2022}. (1) CARLA-Unreal Engine is provided with the head (VR headset) and body (motion capture system) pose. (2) The scenario is generated, including the autonomous vehicles and the digitized pedestrian. (3) The environment is provided to the pedestrian (through VR headset). (4) Autonomous vehicle sensors perceive the environment, including the pedestrian. }
  \label{figure_1}
\end{figure*}

In the following we describe our approach to incorporate real agents behaviors and interactions in CARLA autonomous driving simulator \cite{ref_carla2017} by using immersive virtual reality and human motion capture systems. The idea, schematically represented in Fig. \ref{figure_1}, is to integrate a subject in the simulated scenarios using CARLA and Unreal Engine 4 (UE4), with real time feedback of the pose of his head and body, and including positional sound, attempting to create a virtual experience that is so realistic that the participant feels as though they are physically present in that world and subconsciously accepts it as such (i.e., maximize virtual reality presence). At the same time, the captured pose and motion of the subject is integrated into the virtual scenario by means of an avatar, so that the simulated sensors of the autonomous vehicles (i.e., radar, LiDAR, cameras) can detect their presence as were in the same space. This allows, on the one hand, to obtain synthetic sequences from multiple points of view based on the behavior of real subjects, which can be used to train and test predictive perception models. And on the other hand, they also allow to address different types of interaction studies between autonomous vehicles and real subjects, including external human-machine interfaces (eHMI), under completely controlled circumstances and with absolute safety measures in place.

In this paper, in comparison with our previous work where we already presented the hardware and software architecture \cite{ref_CARLA2022}, we have included the integration of a new motion capture system \cite{ref_XSens2023} and a more detailed description of the computation times and scene processing. Moreover, we have carried out a series of experiments on a novel map and we provide a consistent measure of the sense of \emph{presence} from 18 participants who played the role of a pedestrian in a traffic scenario. Finally, we make some proposals on how to improve the user's immersive experience.

\section{Virtual Reality Immersion Features}
The main goal of our approach is to achieve the total immersion of real pedestrians within a simulator commonly used for autonomous driving testing. We selected CARLA, an open source simulator implemented over UE4 which provides high rendering quality, realistic physics and an ecosystem of interoperable plugins, and we added some features to support an immersive virtual reality system. The user total immersion is achieved through all the functionalities that UE4 presents, along with a virtual reality headset and a set of motion tracking sensors. CARLA is designed as a server-client system, where the server renders the scene and the client generates the agents operating within the dynamic traffic scenario. Communication between the client and the server is done via sockets.

\begin{figure*}[ht]
  \centering
  \includegraphics[width=0.99\linewidth]{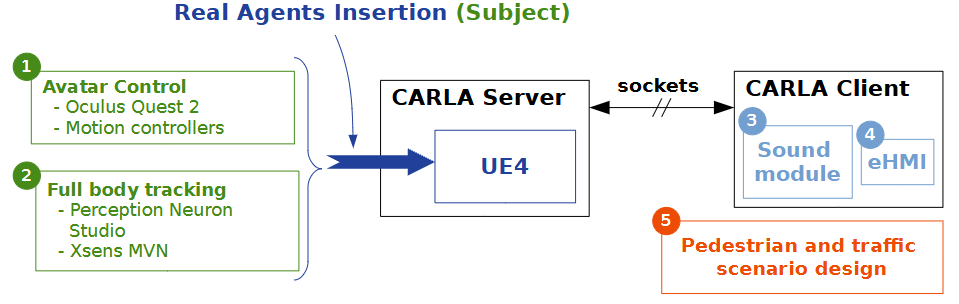}
  \caption{System Block Diagram. Adapted from \cite{ref_CARLA2022}.}
  \label{fig_Block_Diagram}
\end{figure*}

The added features to the simulator for the insertion of real agent behaviours in the CARLA server are based on the five points depicted in Fig. \ref{fig_Block_Diagram}: 1) \textbf{Avatar control}: from the CARLA's blueprint library that collects the architecture of all its actors and attributes, we modify the pedestrian blueprints to create an immersive and maneuverable VR interface between the real agent and the virtual world; 2) \textbf{Body tracking}: we use a set of inertial sensors and proprietary external software to capture the subject's motion through the real scene, and we integrate the avatar's motion into the simulator via \emph{.bvh} files; 3) \textbf{Sound design}: given that CARLA is an audio-less simulator, we incorporate positional sound into the environment to enhance the subject's immersion; 4) \textbf{eHMI integration}: in order to enable communication between autonomous vehicles and other road users to address interaction studies; 5) \textbf{Scenario simulation}: we design traffic scenarios by using the CARLA client, controlling the behaviour of vehicles and other pedestrians.

\subsection{Avatar Control}
CARLA's blueprints (that include sensors, static actors, vehicles and walkers) have been specifically designed to be managed through the Python client API. Vehicles that populate the scenario are actors that incorporate special internal components that simulate the physics of wheeled vehicles and can be driven by functions that provide driving commands (such as throttle, steering or braking). Walkers are operated in the same way and their behavior is directed from the client by a controller, so they are far from adopting behaviors of real pedestrians. 

To support an immersive interface for a real actor, we modify a walker blueprint to make an inverse kinematics setup for full-body scale VR. The tools employed to capture the actor movement are: \textit{a) Oculus Quest 2} (for head tracking and user position control), and \textit{b) Motion controllers} (for both hands tracking). The Oculus Quest 2 safety distance system delimits the playing area through which the subject can move freely. The goal is to allow the subject to move within the established safety zone that purposefully corresponds to a specific area of the CARLA map. 

Firstly, we modify the blueprint by attaching a virtual camera to the head of the walker whose image provided is projected onto the lenses of the VR glasses giving a first-person sensation to the spectator. The displacement and perspective of the walker are also activated, from certain minimum thresholds, with the translation and rotation of the VR headset. The skeletal mesh is another element of the blueprint that we can vary to give the walker another appearance. 

That way, the immersion of a real pedestrian is achieved by implementing a head-mounted display (HMD) and creating an avatar in UE4. The subject wears the VR glasses and also controls the avatar movement throughout the preset area for the experiments.

\subsection{Full-Body Tracking}
On the other hand, head and hand tracking (by mean the VR headset and both motion controllers) serve to adapt the pose of the avatar's neck and hands in real time, but are not enough to represent the full pose of the subject within the simulator. There exit multiple options of motion capture (MoCap) system to do this, including vision-based systems with multiple cameras and inertial measurement units \cite{ref_Menolotto2020}. 

In our case, we have considered the use of two inertial wireless sensor systems: \textit{(i) Perception Neuron Studio (PNS)} motion capture system \cite{ref_PNS2022}, as a compromise solution between accuracy and usability. Each MoCap system includes a set of inertial sensors and straps that can be put on the joints easily, as well as a software for calibrating and capturing precise motion data. \textit{(ii) XSens MVN}, another full body motion analysis system \cite{ref_XSens2023} made up of 17 inertial units (MTw). Based on a biomechanical model, MVN Analyze provides 3D information on joints, center of mass, as well as position, velocity and acceleration parameters for each of the body segments. Both systems allow integration with other 3D rendering and animation software, such as iClone, Blender, Unity or UE4. XSens MVN is a more expensive solution that includes a more sustainable calibration process over time, more exhaustive data processing, and a specific plugin to add the full avatar pose in Unreal Engine in real time.

\subsection{Sound Design}
Since CARLA simulator is world audio absent, the integration of a sound module is another technique to enhance the sensation of \emph{presence} in the virtual world. Sound design and real-world isolation is also essential for interaction with the environment, as humans use spatial sound cues to track the location of other actors and predict their intentions. We incorporate ambient sounds of birds singing and wind, as well as the engines sounds of the vehicles parameterized by its throttle and brake actions. In cases where other pedestrians are involved in the scene, we propose adding other sounds such as conversation or their footsteps so that the subject can be more aware that they are present.

\subsection{External Human-Machine Interfaces (eHMI)}
\begin{figure}[t]
  \centering
  \includegraphics[width=0.8\linewidth]{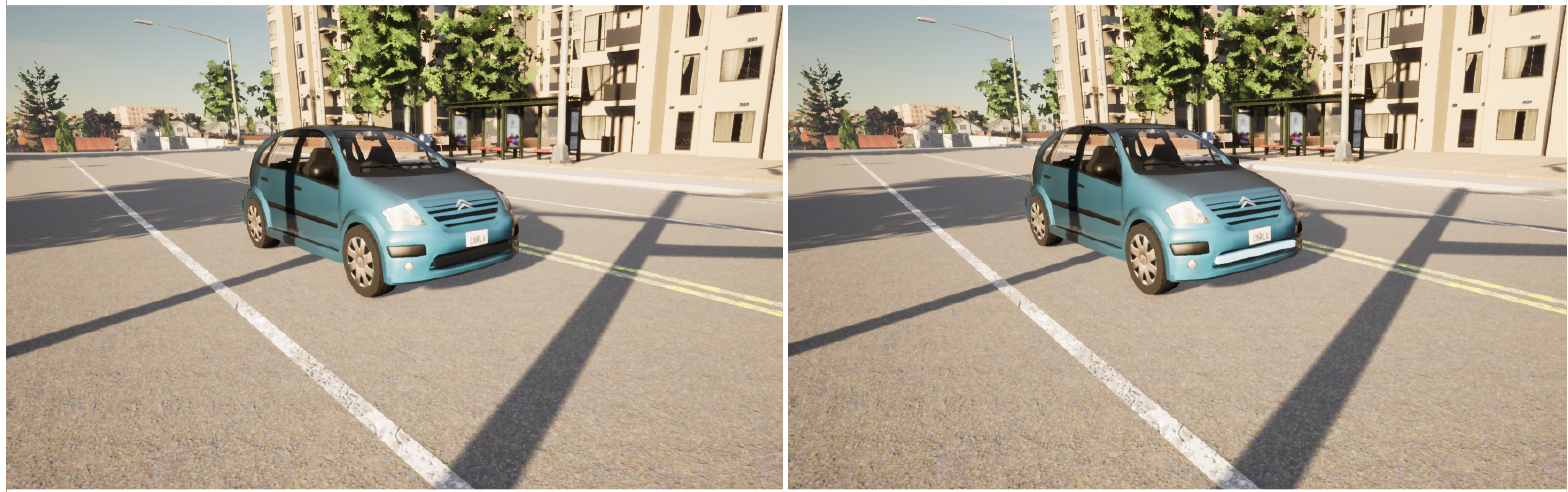}
  \caption{Left: vehicle with eHMI deactivated. Right: vehicle with eHMI activated \cite{ref_CARLA2022}.}
  \label{fig:eHMI}
\end{figure}

In our experiments we include external human-machine interfaces (eHMI) to enable communication between road users. The autonomous vehicles can communicate their status and intentions to the real subject by the proposed eHMI design. As appeared in Fig. \ref{fig:eHMI}, it consists of a light strip along the entire front of the vehicle which changes color depending the information is desired to transmit. This allows studying the influence of the interface on decision making when the pedestrian's trajectory converges with the one followed by the vehicle in the virtual scenario.

\subsection{Traffic Scenario Simulation}
CARLA offers different options to simulate specific traffic scenarios. The Traffic Manager is a module very useful to populate a simulation with realistic urban traffic conditions. Using multiple threads and synchronous messaging, it can propitiate all vehicles to follow certain behaviors (e.g., not exceeding speed limits, ignore traffic light conditions, ignore pedestrians, or force lane changes).

The subject is integrated into the simulator on a map that includes a 3D model of a city. Each map is based on an OpenDRIVE file that describes the fully annotated road layout. This feature allows us to design our own maps as well as implement georeferenced maps taken from the real world. This opens up infinite possibilities for recreating scenarios according to the needs of the study.

\section{System implementation}

\begin{figure*}[ht]
  \centering
  \includegraphics[width=0.99\linewidth]{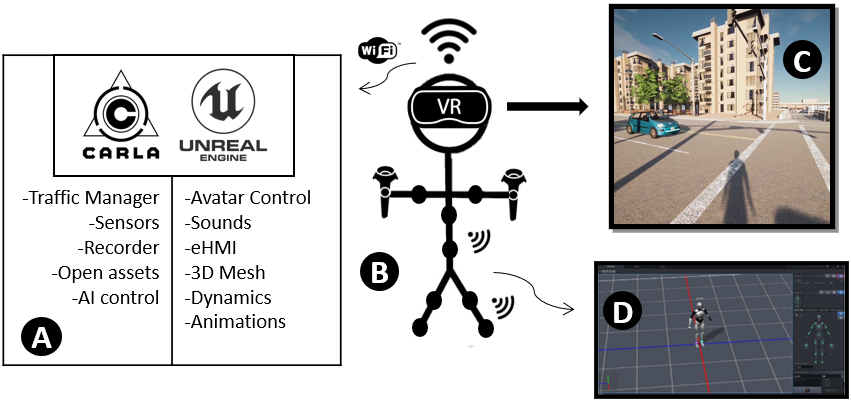}
  \caption{System Schematics \cite{ref_CARLA2022}. (A) Simulator CARLA-UE4. (B) VR headset, motion controllers and body sensors. (C) Spectator View in Virtual Reality. (D) Full-body tracking in Axis Studio or MVN Analyze.}
  \label{figure_2}
\end{figure*}

The overall scheme of the system is shown in Fig. \ref{figure_2}. In the next sections we describe the hardware and software implemented architectures, and the processes of recording and playback of the scenes.

\subsection{Hardware setup}
\begin{figure}[t]
  \centering
  \includegraphics[width=0.8\linewidth]{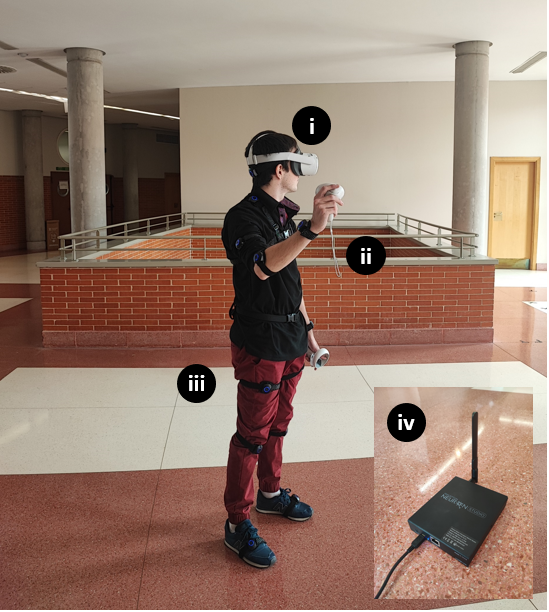}
  \caption{Hardware setup \cite{ref_CARLA2022}. (i) VR headset (Quest 2): transfer the image from the environment to the performer. (ii) Motion controllers: allow control of the avatar's hands. (iii) PN Studio sensors: provide body tracking withstanding magnetic interference. (iv) Studio Transceiver: receives sensors data wirelessly by 2.4GHz.}
  \label{figure_3}
\end{figure}

The complete hardware configuration is depicted in Fig. \ref{figure_3}. We employ the Oculus Quest 2 as our head-mounted device (HMD), created by Meta, which has 6GB RAM processor, two adjustable 1832 x 1920 lenses, 90Hz refresh rate and an internal memory of 256 GB. Quest 2 features WiFi 6, Bluetooth 5.1, and USB Type-C connectivity, SteamVR support and 3D speakers. For full-body tracking we use PNS or XSens solution with inertial trackers. The kit includes standalone VR headset, 2 motion controllers, 17 inertial body sensors, 14 set of straps, 1 charging case and 1 transceiver. During the experiments, we define a preset area wide enough and free of obstacles where the subject can act as a real pedestrian inside the simulator. Quest 2 and motion controllers are connected to PC via Oculus link or WiFi as follows:

\begin{itemize}
\item Wired connection: via the Oculus Link cable or other similar high quality USB 3. 
\item Wireless connection: via WiFi by enabling Air Link from the Meta application, or using Virtual Desktop and SteamVR.
\end{itemize}

The subject puts on the straps of the appropriate length and places the body sensors into the bases. The transceiver is attached to the PC via USB. Quest 2 enables the "VR Preview" in the UE4 editor of the build version of CARLA for Windows.

\subsection{Software setup}
VR Immersion System is currently dependent on UE4.24 and Windows 10 OS due to CARLA build, and Quest 2 Windows-only dependencies. Using TCP socket plugin, all the actor locations and other useful parameters for the editor are sent from the Python API to integrate, for example, the positional sound emitted by each actor and the handling of the eHMI activation of the autonomous vehicle. "VR Preview" projects the game onto the lenses of the HMD. Perception Neuron Studio and XSens MVN work with Axis Studio and MVN Analyze software respectively, supporting up to 3 subjects at a time in the same scene.

\subsection{Recording, Playback and Motion Perception}
When running experiments, certain computational time constraints must be met so that the real subject introduced by virtual reality can perform a natural behavior. The \emph{simulation step} is defined as the time of the scene that is executed at each simulator tick. Under standard conditions, this is not forced to coincide with the \emph{rendering time}, which is the actual time that the architecture takes to process a simulation step. We face the challenge that for the actions of the external agent to be meaningful within the simulated scene, the simulation step and its render time must match.

The rendering time is determined by hardware limitations (i.e., the capacity of the GPU used) and by the number of tasks that are intended to be handled during the simulation. In addition, to attend the immersive sensation, the virtual environment displayed from the VR glasses must show a stable image to the performer so him/her can interact with the world of CARLA. Since the simulated sensors of the autonomous vehicles (i.e. radar, LiDAR, cameras) involve a lot of computations, the scene cannot be reproduced at more than 2 FPS, preventing a successful immersion. To overcome this difficulty, we remove the sensors blueprints, record the simulation data and play it back for later analysis. This allows us to perform the experiments in virtual reality at 18.18 FPS. 

CARLA has a native record and playback system that serializes the world information in each simulator tick for post-simulation recreation. However, this is only intended for tracking actors managed by the Python API and does not include the subject avatar or motion sensors. Along with the recording of the state of the CARLA world, in our case the recording and playback of the complete body motion of the external agent is essential. In our approach we use the Axis Studio or MVN Analyze software to record the body motion during experiments. The recording is exported in a \emph{.bvh} file which is subsequently integrated into the UE4 editor.

Once the action is recorded, the simulation is played back with all the blueprints included since the rendering time does not need to be adjusted to any constraint. Then, the simulated sensors of the autonomous vehicles perceive the skeletal mesh of the avatar and its path followed, as well as the specific pose of all its joints (i.e., body language).

\section{Results}
This section presents the design of some usability examples and an evaluation of the immersive experience provided by the interface for real pedestrians in the CARLA autonomous driving simulator.

\begin{figure}
  \centering
  \includegraphics[width=0.9\linewidth]{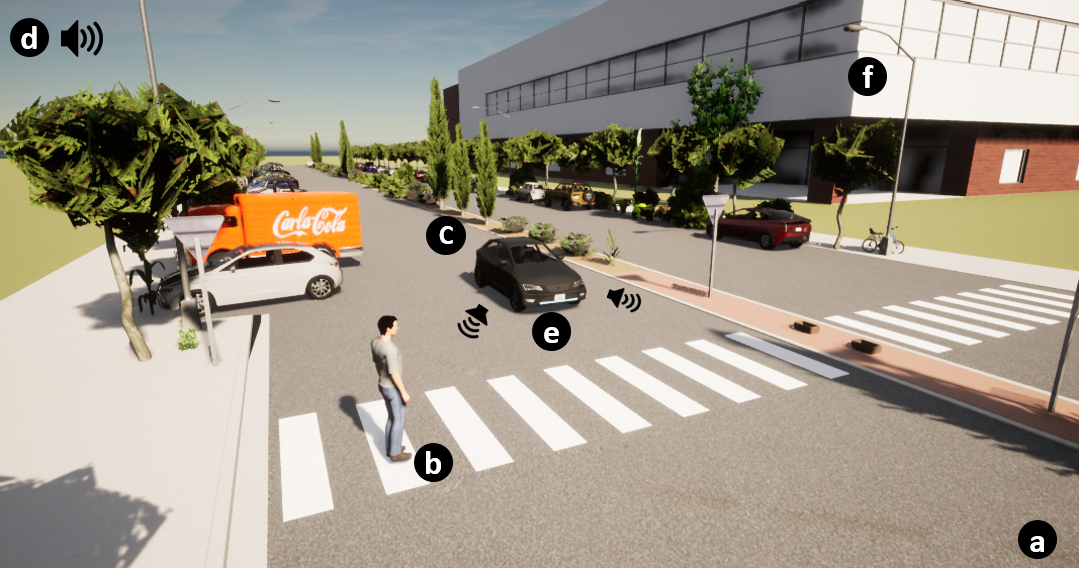}
  \caption{Simulation of Interactive traffic situations. (a) 3D world design. (b) Pedestrian matches the performer avatar. (c) Autonomous vehicle. (d) Environment sounds and agents sounds. (e) eHMI. (f) Street lighting and traffic signs.}
  \label{figure_4}
\end{figure}

\subsection{Usability examples}
To attend our purposes, the implemented traffic scenario (depicted in Fig. \ref{figure_4}) must propitiate interactions between autonomous vehicles and the user of the virtual reality glasses and motion capture system who walks through the environment as a pedestrian \cite{ref_Martin2023a}. The first step is to select a suitable map where to develop the action. We downloaded the map data of the university area from OpenStreetMap \cite{ref_openstreetmap2023} and converted it to an OpenDRIVE format which can be ingested into CARLA. This allows us to obtain the geometry information of a real pedestrian crossing and replicate its same visibility conditions. 

\begin{figure}
  \centering
  \includegraphics[width=0.9\linewidth]{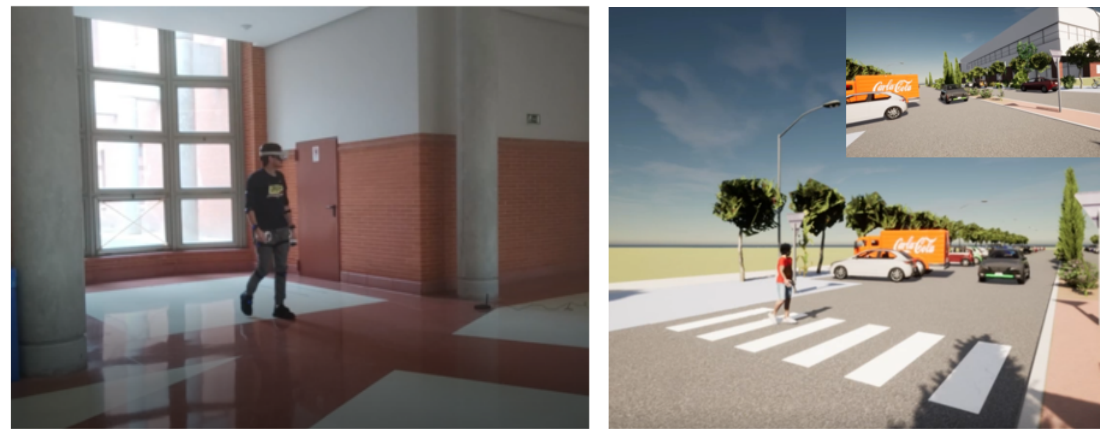}
  \caption{Left: real scenario, VR and motion capture setting. Right: simulated scenario and pedestrian's view.}
  \label{figure_r}
\end{figure}

\begin{figure}
  \centering
  \includegraphics[width=\linewidth]{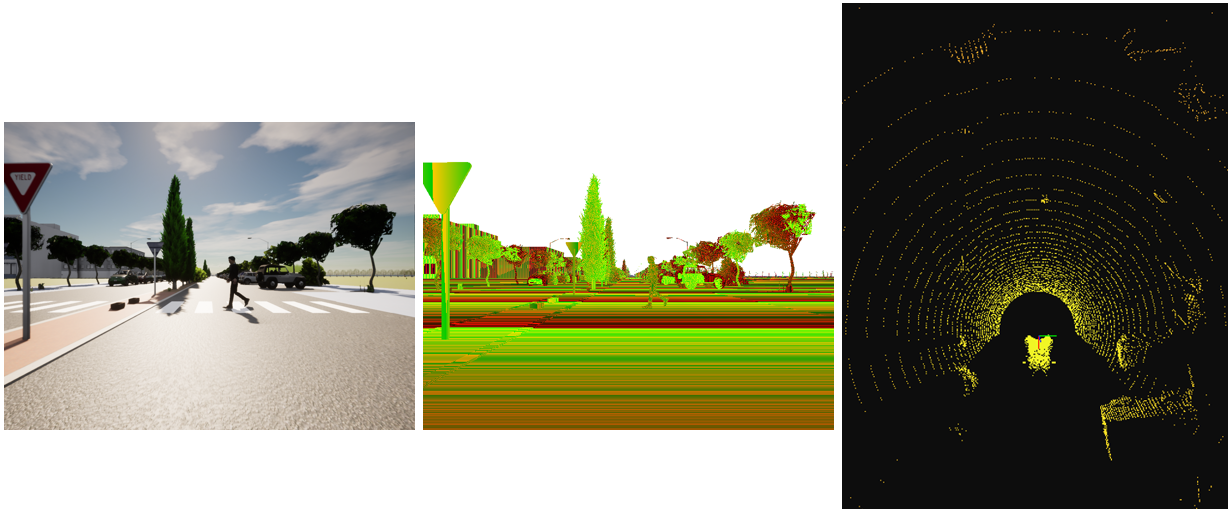}
  \caption{Virtual sensors output: cameras (RGB, depth), and LiDAR point cloud (ray-casting).}
  \label{figure_6}
\end{figure}

When running the scene (see Fig. \ref{figure_r}), an autonomous vehicle circulates on the road when reaches the pedestrian crossing. The pedestrian on the edge of the sidewalk is ordered to cross the road when they consider it safe, and receives information on status and vehicle intentions through an eHMI. In addition, the pedestrian can hear the engine of the vehicle approaching, which can influence the decision to cross sooner or later. From the CARLA client, it is possible to pre-program the behavior of the autonomous vehicle so that it ignores the pedestrian and does not stop or performs a braking maneuver and gives way. To observe its impact on the pedestrian's attitude (i.e., on the interaction), more or less aggressive braking maneuvers can be applied, and the external HMI can be activated or deactivated. Lighting and weather conditions are also adjustable. Sensors attached to the vehicle capture the image of the scenario and detect the pedestrian, as shown in Fig. \ref{figure_6}.

\subsection{User experience evaluation}
A sample of 18 experimental participants, consisted of 12 male and 6 female who ranged in age from 24 to 62, were instructed to take part in the scene in the role of the pedestrian and completed a 15-item presence scale (depicted in Appendix A) to asses the quality of immersion. \emph{Self-presence} examines how much a user extends features of their identity into a virtual world while represented by an avatar. \emph{Autonomous vehicle} and \emph{environmental presence} measure how a user treats actors and environments in mediated space as if they were real. In addition, we request participants for open comments about their performance.

As a means of assessing the test's reliability, we compute Cronbach's Afla (\begin{math}\alpha\end{math}= .707) which indicates an acceptable internal consistency. Most of the participants felt a strong self-presence (M=4.04, SD=.953) perceiving the displacement and hands of the avatar as their own. Regarding autonomous vehicle presence (M=3.94, SD=.967), the engine noise was the main point of contention among the participants, as some found it highly helpful in identifying the vehicle, while others either didn't notice it or found it irritating. Environmental-presence (M=4.34, SD=.627) got the highest score; the participants stated the appearance of the environment was that of a real crosswalk.

Self-presence and environmental presence were satisfactory, while most feedback was directed at improving the presence of the autonomous vehicle. Its braking maneuver did not feel threatening in the sense that was appreciated too conservative, and the vehicle dynamics did not help to anticipate the point at which it was going to stop.

\section{Conclusions and Future Work}
We have developed a framework to enable real-time interaction between real agents and simulated environments. The initial focus is on the integration of pedestrians in traffic scenarios, for which a virtual reality interface has been implemented in the CARLA simulator for autonomous driving. The virtual world is displayed on the glasses lenses at 18.18 FPS. The performer pose is registered by a motion capture system, generating useful sequences to train and validate predictive models to, for example, predict future actions and trajectories of traffic agents. This paper has presented some possibilities and usability cases that this system can address.

As future works, it is intended to improve some aspects of the immersive experience. XSens MVN will replace the PNS system to represent the user's full body on the avatar in real time. We will apply improvements in the dynamics of the vehicle (e.g., an inclination of its frontal part at the moment of its stop). The addition of other agents on the scene, such as vehicles traveling in the other direction, and other pedestrians, will be considered to enable different types of interaction studies. Furthermore, one of our main goals is to provide a measure of the \emph{behavioral gap} by replicating interaction and communication studies in equivalent real and virtual environments.

\subsubsection{Acknowledgements} This work was funded by Research Grants PID2020-114924RB-I00 and PDC2021-121324-I00 (Spanish Ministry of Science and
Innovation) and partially by S2018/EMT-4362 SEGVAUTO
4.0-CM (Community of Madrid).  D. Fernández Llorca acknowledges funding
from the HUMAINT project by the Directorate-General
Joint Research Centre of the European Commission.

\section*{Disclaimer}
The views expressed in this article are purely those of the authors and may not, under any circumstances, be regarded as an official position of the European Commission. 

%
%
%

\begin{thebibliography}{8}

\bibitem{ref_Kalra2016}
Kalra, N., Paddock, S. M.: Driving to safety: how many miles of driving would it take to demonstrate autonomous vehicle reliability? RAND Corporation, Research report, 2016. 

\bibitem{ref_Llorca2023}
Fernández-Llorca, D., Gómez, E.: Trustworthy Artificial Intelligence Requirements in the Autonomous Driving Domain. Computer, vol. 56, no. 2, pp. 29-39, 2023, doi: 10.1109/MC.2022.3212091.

\bibitem{ref_Izquierdo2022}
Izquierdo Gonzalo, R., Salinas Maldonado, C., Alonso Ruiz, J., Parra Alonso, I., Fernández Llorca, D., Sotelo, M. Á.: Testing Predictive Automated Driving Systems: Lessons Learned and Future Recommendations.  IEEE Intelligent Transportation Systems Magazine, vol. 14, no. 6, pp. 77-93, 2022, doi: 10.1109/MITS.2022.3170649.

\bibitem{ref_Stocco2022}
Stocco, A., Pulfer, B., Tonella, P.: Mind the Gap! A Study on the Transferability of Virtual vs Physical-world Testing of Autonomous Driving Systems. IEEE Transactions on Software Engineering, \doi{10.1109/TSE.2022.3202311}, pp. 1--13, 2022.

\bibitem{ref_GarciaDaza2022}
García Daza, I., Izquierdo, R., Martínez, L. M., Benderius, O., Fernández Llorca, D.: Sim-to-real transfer and reality gap modeling in model predictive control for autonomous driving. Applied Intelligence, 2022, \doi{10.1007/s10489-022-04148-1}.

\bibitem{ref_Eady2019}
Eady, T.: (2019, May) Simulations can’t solve autonomous driving because they lack important knowledge about the real world – large-scale real world data is the only way. \url{https://medium.com/@trenteady/simulation-cant-solve-autonomous-driving-because-it-lacks-necessary-empirical-knowledge-403feeec15e0}, Last accessed April 8, 2022.

\bibitem{ref_carla2017}
Dosovitskiy, A., Ros, G., Codevilla, F., Lopez, A., Koltun, V.: {CARLA}: {An} Open Urban Driving Simulator. Proceedings of the 1st Annual Conference on Robot Learning, pp. 1--16, 2017.

\bibitem{ref_CARLA2022}
Mart{\'{\i}}n Serrano, S., Fern{\'{a}}ndez Llorca, D., Garc{\'{\i}}a Daza, I., Sotelo, M. A.: Insertion of Real Agents Behaviors in {CARLA} Autonomous Driving Simulator. Proceedings of the 6th International Conference on Computer-Human Interaction Research and Applications, {CHIRA 2022}, pp. 23--31. \doi{10.5220/0011352400003323}

\bibitem{ref_Menolotto2020}
Menolotto, M., Komaris, D., Tedesco, S., O’Flynn, B., Walsh, M.: Motion Capture Technology in Industrial Applications: A Systematic Review. Sensors, 20(19), 2020. \doi{10.3390/s20195687}

\bibitem{ref_PNS2022}
Noitom (2022), \url{https://neuronmocap.com/perception-neuron-studio-system}. Last accessed March 22, 2023

\bibitem{ref_XSens2023}
Movard (2023), \url{https://www.movard.es/productos/xsens/}. Last accessed March 22, 2023

\bibitem{ref_Martin2023a}
Martín Serrano, S., Izquierdo, R., García Daza, I., Sotelo, M. Á., Fernández Llorca, D.: Digital twin in virtual reality for human-vehicle interactions in the context of autonomous driving. arXiv:2303.11463, 2023. 

\bibitem{ref_openstreetmap2023}
OpenStreetMap, \url{https://www.openstreetmap.org/about}. Last accessed March 23, 2023








\end{thebibliography}
%

\section*{Appendix A}

\subsection*{Self-presence Scale items}

To what extent did you feel that… (1= not at all – 5 very strongly)

\begin{enumerate}
\item You could move the avatar's hands.
\item The avatar's displacement was your own displacement.
\item The avatar's body was your own body.
\item If something happened to the avatar, it was happening to you.
\item The avatar was you.
\end{enumerate}

\subsection*{Autonomous vehicle presence Scale items} 

To what extent did you feel that… (1= not at all – 5 very strongly)

\begin{enumerate}
\item The vehicle was present.
\item The vehicle dynamics and its movement were natural.
\item The sound of the vehicle helped you to locate it.
\item The vehicle was aware of your presence.
\item The vehicle was real.
\end{enumerate}

\subsection*{Environmental presence Scale items} 

To what extent did you feel that… (1= not at all – 5 very strongly)

\begin{enumerate}
\item You were really in front of a pedestrian crossing.
\item The road signs and traffic lights were real.
\item You really crossed the pedestrian crossing.
\item The urban environment seemed like the real world.
\item It could reach out and touch the objects in the urban environment.
\end{enumerate}

\end{document}